\documentclass[pmlr,twocolumn,10pt]{jmlr} %

\usepackage{setspace}

\usepackage{booktabs}
\usepackage{multirow}
\usepackage[load-configurations=version-1]{siunitx} %

\usepackage{tikz}
\usetikzlibrary{positioning,arrows.meta,quotes}
\usetikzlibrary{shapes,snakes}
\usetikzlibrary{bayesnet}
\tikzset{>=latex}
\tikzstyle{plate caption} = [caption, node distance=0, inner sep=0pt,
below left=5pt and 0pt of #1.south]

\usepackage{bbm}
\newcommand\pb[1]{\ensuremath{\left[ #1 \right]}} %
\newcommand\pc[1]{\ensuremath{\left\{ #1 \right\}}} %

\newcommand\E{\ensuremath{\mathbb{E}}}
\newcommand\R{\ensuremath{\mathbb{R}}} %

\newcommand\sR{\ensuremath{\mathcal{R}}} 
\newcommand\sD{\ensuremath{\mathcal{D}}} 
\newcommand\sI{\ensuremath{\mathcal{I}}}

\newcommand\bbP{\ensuremath{\mathbbm{P}}}

\DeclareMathOperator*{\argmax}{arg\,max}

\definecolor{nice_blue}{RGB}{65, 105, 225}
\definecolor{nice_red}{RGB}{168, 34, 34}

\theorembodyfont{\upshape}
\theoremheaderfont{\scshape}
\theorempostheader{:}
\theoremsep{\newline}

\jmlrvolume{LEAVE UNSET}
\jmlryear{2021}
\jmlrsubmitted{LEAVE UNSET}
\jmlrpublished{LEAVE UNSET}
\jmlrworkshop{Machine Learning for Health (ML4H) 2021} %

\title[Image Classification with Consistent Supporting Evidence]{Image Classification with Consistent Supporting Evidence}

 \author{%
  \Name{Peiqi Wang} \Email{wpq@mit.edu}\\
  \Name{Ruizhi Liao} \Email{ruizhi@csail.mit.edu}\\
  \Name{Daniel Moyer} \Email{dmoyer@csail.mit.edu}\\
  \addr Massachusetts Institute of Technology, Cambridge, MA, USA \AND
  \Name{Seth Berkowitz} \Email{sberkowi@bidmc.harvard.edu}\\
  \Name{Steven Horng} \Email{shorng@bidmc.harvard.edu}\\
  \addr Beth Israel Deaconess Medical Center / Harvard Medical School, Boston, MA, USA \AND
  \Name{Polina Golland} \Email{polina@csail.mit.edu} \\
  \addr Massachusetts Institute of Technology, Cambridge, MA, USA
 }

\begin{document}

\maketitle

\begin{abstract}
    Adoption of machine learning models in healthcare requires end users' trust in the system. Models that provide additional supportive evidence for their predictions promise to facilitate adoption. We define consistent evidence to be both compatible and sufficient with respect to model predictions. We propose measures of model inconsistency and regularizers that promote more consistent evidence. We demonstrate our ideas in the context of edema severity grading from chest radiographs. We demonstrate empirically that consistent models provide competitive performance while supporting interpretation.
\end{abstract}

\section{Introduction}
\label{sec:intro}

Identifying radiological findings and inferring disease stages from medical images is common in clinical practice. Many models make predictions without explaining the conclusion. In contrast, human experts often provide specific explanation based on prior knowledge of human physiology to support their image-based diagnosis. We aim to build models that are transparent in the reasoning process, at an appropriate level of understanding consumable by end users, e.g., clinicians. What additional information should a machine learning model provide to gain the trust of its end users? We propose a solution motivated by an example of how radiologists themselves operate.

\begin{figure}[ht]
\floatconts
    {fig:teaser_consistent_evidence}
    {\caption{Our model provides prediction of disease stage $y$ and supporting evidence. We use $\sI(c)$ to denote the set of evidence labels detected in the image that directly support disease stage $c$. We show examples of inconsistent evidence highlighted in {\color{nice_red} red} produced by the baseline model (left column). Our proposed regularizer corrects these mistakes so that predicted evidence label becomes compatible (top right) and sufficient (bottom right).}}
    {%
    \includegraphics[width=\linewidth]{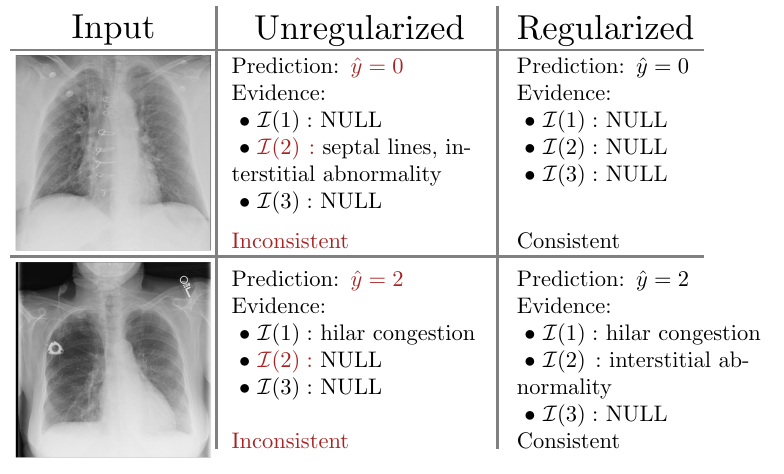}%

    }
\end{figure}

Radiological \textit{findings} are concepts determined as useful by radiologists. Findings include image features, pathological states, and observations about the underlying physiology \citep{glueckerClinicalRadiologicFeatures1999}. The radiologists aggregate the findings to provide an overall interpretation of the image. They support the eventual diagnosis by providing an account of the identified findings based on prior knowledge of relationships between findings and the patient's physiological state. We propose and demonstrate an approach that re-capitulates the reasoning process of domain experts. In addition to primary predictions, the model provides supporting evidence, i.e., findings, deemed useful by the end users.

It is critical that predictions and supporting evidence are consistent with each other. In practice, radiologists cannot draw their conclusions based on incompatible evidence, nor could they support their conclusions with insufficient evidence. Similarly, end users will question the credibility of a model when its predictions and accompanying evidence are incompatible or insufficient.

In this paper, we build explainable models that supplement their predictions with consistent supporting evidence, illustrated in Figure \ref{fig:teaser_consistent_evidence}. We define measures of inconsistency between the model's primary output and its supporting evidence and propose simple regularizers that encourage the classifier to be more consistent. We demonstrate that we can train consistent models without loss in performance  in the context of pathology grading from a chest radiograph.

\section{Related Work}

\subsection{Interpretable Machine Learning}

While model interpretability is an important topic in machine learning, few methods take the end users' needs into account. For example, some method localize image regions important for a prediction \citep{zhouLearningDeepFeatures2016,selvarajuGradCAMVisualExplanations2016}, but fail to express what properties of the image region are associated with the model output. Others aim to use a simpler model \citep{caruanaIntelligibleModelsHealthCare2015} or to approximate the behavior of a complex model with a simpler one \citep{ribeiroWhyShouldTrust2016}. While effective for handling low dimensional tabular data where the covariates are physically meaningful attributes, such methods are less useful for extremely high dimensional imaging data. Our work provide clinically meaningful supporting evidence useful to end users of the system, rather than support the developers' understanding of how the model reaches its decision.

Another approach is to train a classifier whose predictions rely on higher-level concepts. Unsupervised methods can make for a more interpretable model for general purpose tasks but cannot take advantage of strong domain knowledge ubiquitous in healthcare \citep{alvarezmelisRobustInterpretabilitySelfExplaining2018}. Alternatively, concept bottleneck models that learn concepts with supervision have been applied to arthritis grading \citep{kohConceptBottleneckModels2020}, retinal disease classification \citep{defauwClinicallyApplicableDeep2018}, and other applications \citep{loschInterpretabilityClassificationOutput2019,bucherSemanticBottleneckComputer2019}. This strategy relies on the appropriate choice of the concepts to maintain good performance. In contrast, our model predictions do not solely rely on supporting evidence, thus avoiding the undue influence the choice of supporting evidence or concepts might have on the accuracy of the task label prediction.

In our work, predictions and evidence are separate outputs of the model. Most closely related prior method focus on learning a mapping to the product space of the task label and supporting evidence \citep{hindTEDTeachingAI2019,codellaTeachingAIExplain2019}, with application to text classification \citep{zaidanUsingAnnotatorRationales2007,zhangRationaleAugmentedConvolutionalNeural2016}. In contrast, we learn a structured output where known relationships between predictions and evidence are enforced. We inject domain specific knowledge and require our model to provide supporting evidence that is clinically feasible under a specific prediction. 

A recently demonstrated unsupervised strategy that requires a forward model that relates supporting evidence to a subset of the input features is also relevant \citep{raghuLearningPredictSupporting2021}. This method can be difficult to implement in our radiograph grading task as it assumes knowledge of an accurate forward model, from supporting evidence to a high dimensional image, which is infeasible for most medical imaging problems.

\subsection{Logical Constraints}

There are multiple ways to represent symbolic constraints. For example, trees have been used to express subsumption relationships between attributes, e.g., hierarchical annotation of medical images \citep{dimitrovskiHierarchicalAnnotationMedical2011}. Hierarchical multi-label learning aims to enforce such constraints \citep{biMultiLabelClassificationTree2011,yanHDCNNHierarchicalDeep2015,wehrmannHierarchicalMultiLabelClassification2018,giunchigliaCoherentHierarchicalMultiLabel2020}. Unfortunately, trees are overly restrictive and cannot express consistency constraints that are important in our application. 

Alternatively, boolean statements can be quite expressive in representing logical constraints. Logical constraints on model outputs can be enforced by replacing logical operators with their subdifferentiable fuzzy t-norms \citep{diligentiIntegratingPriorKnowledge2017,liLogicDrivenFrameworkConsistency2019} or through the use of specialized loss functions \citep{xuSemanticLossFunction2018}. Our approach to representing and enforcing logical constraints is both easier to interpret and simpler to implement. Moreover, we enforce logical constraints by making changes to how our model provides supporting evidence while keeping main task predictions unchanged.

\section{Method}
\label{sec:method}

In this section, we define consistency of evidence and introduce measures of inconsistency. We provide an example application to ground our definitions. We construct novel loss functions that encourage consistency and discuss optimization that arises when training classifiers with consistent supporting evidence.

\subsection{Problem Setup}
\label{sec:problem_setup}

Let $x,y$, and $z=(z_1,\ldots,z_K)$ be random variables representing an image, a $C$-class task label, and $K$ binary evidence labels. Let $\sD_t$ be a data set that includes pairs $(x,y)$ and $(x,z_k)$ for $k=1,\ldots,K$. In this work, we do not assume full tuples $(x,y,z_1,\ldots,z_K)$ are available. Learning joint predictors from available sub-tuples is an interesting direction that is outside the scope of this paper. Moreover, we allow the same image $x$ to be included as part of several different pairs in the data set $\sD_t$.

We use the training data set $\sD_t$ to build probabilistic classifiers $p(y \mid x;\theta)$ and $p(z_k \mid x; \theta)$ for $k=1,\ldots,K$. The maximum a posteriori (MAP) estimates of the task label $y$ and evidence labels $z$ is obtained via
\begin{align}
    \hat{y} 
        &=\,\, \argmax_{c\in\pb{C}} \,\,\, p(y = c \mid x ;\theta), \\
    \hat{z}_k
        &= \argmax_{z\in\pc{-1,+1}} p(z_k = z \mid x; \theta),
        \quad k=1,\ldots,K.
\end{align} 
We use $\hat{y}(x)$ to express explicit dependence of the predicted label $\hat{y}$ on the input image $x$.

\subsection{Consistent Evidence}
\label{sec:consistent_evidence}

We assume that domain experts provide domain specific knowledge in the form of logical constraints between the task label $y$ and the evidence labels $z$. We identify two major logical constraints that are important in our application, specifically that supporting evidence should be compatible and sufficient with the task label.

Let $\sI_1: \pb{C}\to \mathcal{P}(\pb{K})$ be the indexing function for evidence that is incompatible with a particular value of task label, where we use $\mathcal{P}(\cdot)$ to denote the power set. Specifically, if evidence labels $\pc{z_{i_1},\ldots,z_{i_M}}$ are incompatible with task label $y=c$, then $\sI_1(c) = \pc{i_1,\ldots,i_M}$. Let $\sI_2: \pb{C}\to\mathcal{P}(\pb{K})$ be the indexing function for evidence that directly supports a particular value of task label. We assume that $\sI_1$ and $\sI_2$ are provided by domain experts. 
\begin{definition}
    \label{def:consistent_evidence}
    (Consistent Evidence) The task label $y\in\pb{C}$ and the evidence label vector $z=(z_1,\ldots,z_K)\in \{-1,+1\}^K$ are consistent if
    \begin{align}
        &\forall k\in \sI_1(y):\; z_k = -1, \\
        &\exists k\in \sI_2(y):\; z_k = +1.
    \end{align}
\end{definition}
The first criterion specifies that no evidence is incompatible with the task label $y$. The second criterion specifies that there should be at least one direct evidence label present that supports the task label $y$.

In reality, perfectly consistent evidence may not be necessary or possible. For example, domain experts often specify constraints with a notion of uncertainty, e.g., $\sI_1(y)$ is incompatible with $y$ most of the time except for occasional corner cases. In addition, certain direct evidence might be so rare that it becomes impossible to include it in $\sI_2(y)$. Therefore, it is perfectly sensible that there is no direct evidence present in some cases, if we have not included the corresponding evidence label in the construction. This motivates us to consider these constraints in probabilistic terms.

Definition  \ref{def:consistent_evidence} is a specification over the values that random variables can take. The same definition applies to the true data distribution $\pc{y,z}$ and to the predicted distribution $\pc{\hat{y},\hat{z}}$. In practice, we construct training data $\pc{y,z}$ to be perfectly consistent and demonstrate a training method that encourages the model outputs $\pc{\hat{y},\hat{z}}$ to be consistent as well.

Moreover, we are not restricted to predicting the findings in $\cup_{c\in\pb{C}}\sI_2(c)$. If some findings provide useful information but are not directly supportive, they can still be included in the set of evidence labels.

\subsection{Example Application}
\label{sec:example_application}

This section illustrates the construction of indexing functions $\sI_1$ and $\sI_2$ for the pulmonary edema grading task that motivated our work.

Pulmonary edema is defined as an abnormal accumulation of fluid in the lungs. Higher hydrostatic pressure in the vasculature causes more severe symptoms. Typically, radiologists grade the severity of edema based on findings that are typical of the most severe stage of pulmonary edema \citep{glueckerClinicalRadiologicFeatures1999}. 

We use a categorization that identifies four edema severity levels, in order of increasing severity: \textit{no edema} (0), \textit{mild edema} (1), \textit{moderate edema} (2), and \textit{severe edema} (3) \citep{liaoSemisupervisedLearningQuantification2019,horngDeepLearningQuantify2021}. The edema severity grading task involves assigning a severity level $y\in \pc{0,1,2,3}$ to a test image. In this task, there are 4 classes, i.e., $C=4$.

\begin{table}[!ht]
\floatconts
    {tab:severity_evidence}
    {\caption{Findings that directly support a particular severity level.}}
    {\begin{tabular}{ll}
    \toprule
    \bfseries Severity $y$ & \bfseries Findings $\sI_2(y)$ \\
    \midrule
    \multirow{1}{*}{0 (none)}
        & - \\
    \cmidrule{2-2}
    \multirow{3}{*}{1 (mild)}
        & vascular congestion \\
        & hilar congestion \\
        & peribronchial cuffing \\
    \cmidrule{2-2}
    \multirow{2}{*}{2 (moderate)}
        & septal lines \\
        & interstitial abnormality \\
    \cmidrule{2-2}
    \multirow{2}{*}{3 (severe)}
        & air bronchograms \\
        & parenchymal opacity \\
    \bottomrule
    \end{tabular}}
\end{table}

\begin{table}[!h]%
\floatconts
    {tab:example_consistent}
    {\caption{Examples of consistent evidence.\qquad\qquad\qquad}}
    {
        \begin{tabular}{cl}
        \toprule
        \bfseries Severity $y$ & \bfseries Evidence $z$ \\
        \midrule
        \multirow{1}{*}{1}
            & peribronchial cuffing \\
        \cmidrule{2-2}
        \multirow{3}{*}{2}
            & vascular congestion \\
            & septal lines \\
            & interstitial abnormality \\
        \bottomrule
        \end{tabular}
    }
\end{table}

\begin{table}[!h]%
\floatconts
    {tab:example_inconsistent}
    {\caption{Examples of inconsistent evidence. First two examples are incompatible. The latter two examples are insufficient.}}
    {
        \begin{tabular}{cl}
        \toprule
        \bfseries Severity $y$ & \bfseries Evidence $z$ \\
        \midrule
        \multirow{2}{*}{1}
            & hilar congestion \\
            & septal lines \\
        \cmidrule{2-2}
        \multirow{3}{*}{2}
            & vascular congestion \\
            & interstitial abnormality \\ 
            & air bronchograms \\
        \midrule
        \multirow{1}{*}{1}
            & – \\
        \cmidrule{2-2}
        \multirow{1}{*}{3}
            & septal lines \\ 
        \bottomrule
        \end{tabular}
    }
\end{table}
    
In our work, we identify $K=7$ supporting evidence labels deemed useful by clinicians, as shown in Table \ref{tab:severity_evidence}. They are canonical radiological manifestation of the underlying pathology. End users expect presence of these findings to be indicative of a specific edema severity level.

As an example, radiologists grade an image as \textit{moderate edema} if they observe \textit{septal lines} (short parallel lines at the periphery of the lung) or \textit{interstitial abnormality} (excess fluids in the supporting tissue within the lung). Note that presence of evidence from a lower value of edema severity is not inconsistent. For example, radiologists may at the same time observe presence of \textit{vascular congestion} (enlargement of pulmonary veins) and \textit{septal lines} in a moderate edema case.

In the severity grading task, we consider an evidence label as incompatible if its presence directly supports a higher level severity level. Thus define
\begin{align}
    \sI_1(c) = \bigcup_{c'>c}\, \sI_2(c').
\end{align}
As an example, a model that grades an image as \textit{moderate edema} should not use \textit{air bronchograms} (opacification of alveoli) as supporting evidence.

We consider evidence as insufficient when no direct evidence for edema severity grading is present. As an example, a model which grades an image as \textit{severe edema} cannot rely on \textit{septal lines} only to support its prediction.

Tables \ref{tab:example_consistent} and \ref{tab:example_inconsistent} illustrate further examples of consistent and inconsistent evidence, respectively.

\subsection{Measuring Inconsistency}
\label{sec:measure_inconsistency}

We quantify the inconsistency probabilistically based on Definition \ref{def:consistent_evidence}. First, we define a measure of incompatibility as the probability that there is an incompatible evidence label
\begin{align}
    \bbP\pb{
        \bigcup_{k\in \sI_1(y)} \pc{z_k = +1}
    }.
    \label{eq:inconsistency_measure_incompatible}
\end{align}
To facilitate computation, we upper bound this probability using union bound by
\begin{align}
    \sR_1(y,z)
        = \sum_{k\in \sI_1(y)} \bbP\pb{z_k = +1}.
    \label{eq:inconsistency_measure_incompatible_upper_bound_one}
\end{align}
We provide an estimate of incompatibility over data set $\sD$ by taking expectation over its empirical distribution
\begin{align}
    \sR_1(\sD) %
        &= \E_{(y,z)\sim \sD} \pb{
            \sum_{k\in \sI_1(y)} \mathbbm{1}\pb{ z_k = +1 }
        },
    \label{eq:inconsistency_measure_incompatible_upper_bound_dataset} 
\end{align}
where we have replaced $\bbP\pb{z_k=+1}$ with $\mathbbm{1}\pb{z_k = +1}$ since $z_k$ is binary valued. Intuitively, $\sR_1(\sD)$ is the average count of evidence labels incompatible with the task label.

In this work, we use Equation \ref{eq:inconsistency_measure_incompatible_upper_bound_dataset} as a measure of incompatible evidence. We also note that one could eliminate the dependency on the size of $\sI_1(y)$ by defining
\begin{align}
    \E_{(y,z)\sim \sD} \pb{
            \frac{1}{\left| \sI_1(y) \right|} \sum_{k\in \sI_1(y)} \mathbbm{1}\pb{ z_k = +1 }
        }.
    \label{eq:inconsistency_measure_incompatible_upper_bound_dataset_normalized} 
\end{align}
We prefer the measure of incompatibility defined in Equation \ref{eq:inconsistency_measure_incompatible_upper_bound_dataset} for measuring incompatibility as it arises naturally from bounding the probability of incompatible evidence.

Similarly, we define a measure of insufficiency as the probability that there is no sufficient evidence
\begin{align}
    \bbP\pb{
        \bigcap_{k\in \sI_2(y)} \pc{z_k = -1}
    },
    \label{eq:inconsistency_measure_insufficient}
\end{align}
which leads to an upper bound
\begin{align}
    \sR_2(y,z)
        &= \min_{k\in \sI_2(y)} \bbP\pb{z_k = -1}
        \label{eq:inconsistency_measure_insufficiency_upper_bound_one}
\end{align}
and its empirical estimate
\begin{align}
    \sR_2(\sD)
        &= \E_{(y,z)\sim \sD} \pb{
            \min_{k\in \sI_2(y)} \pb{ 1 - \mathbbm{1}\pb{ z_k = +1 }  }
        } \\
        &= 1 - \E_{(y,z)\sim \sD} \pb{
            \max_{k\in\sI_2(y)} \mathbbm{1}\pb{z_k = +1}
        }.
        \label{eq:inconsistency_measure_insufficiency_upper_bound_dataset} 
\end{align}
Note that $\sR_2(\sD)$ is the average count of absence of direct evidence.

Now we can provide an upper bound on probability of inconsistent evidence $\sR(y,z) = \sR_1(y,z) + \sR_2(y,z)$ and its empirical estimate $\sR(\sD) = \sR_1(\sD) + \sR_2(\sD)$.

\subsection{Consistency Regularization}
\label{sec:consistency_regularization}

Models trained naively to predict labels $y$ and $z$ jointly are not guaranteed to be consistent. Here, we provide regularizers that encourage supporting evidence to be more consistent.

Observe that Equations \ref{eq:inconsistency_measure_incompatible_upper_bound_one} and \ref{eq:inconsistency_measure_insufficiency_upper_bound_one} are upper bounds on the true probability of model being inconsistent. We can simply use these upper bounds, or modification thereof, as regularizers. We opt to use cross entropy to avoid inconsistent evidence.

To penalize incompatibility, we define
\begin{align}
    \sR_1(\theta)
        &= - \E_{x\sim \sD} \pb{
            \sum_{k\in \sI_1(\hat{y}(x))}
            \ln p(z_k = -1 \mid x; \theta)
        }.
    \label{eq:inconsistency_r1}
\end{align}
Intuitively, $\sR_1(\theta)$ penalizes evidence probability that is incompatible with the predicted task label. Including $\sR_1(\theta)$ in the loss function is equivalent to supplying pseudo negative samples for evidence obtained from the predicted task label $\hat{y}$. 

Instead of penalizing incompatibility with respect to MAP estimate of the task label $\hat{y}(x)$, we can penalize incompatibility for each value of task label weighted by the posterior probability, i.e.,
\begin{align}
    \widetilde{\sR}_1(\theta) = -\E_{x\sim \sD} \Bigg[ \Bigg.
        \sum_{c\in[C]} \sum_{k\in \sI_1(c)} &p(y=c\mid x;\theta) \, \cdot  \notag \\
        & \ln p(z_k = -1 \mid x;\theta) 
    \Bigg. \Bigg].
    \label{eq:inconsistency_r1_soft}
\end{align}
In contrast to Equation \ref{eq:inconsistency_r1} where gradients cannot flow through $\hat{y}(x)$ due to the $\argmax$ operator, Equation \ref{eq:inconsistency_r1_soft} provides a softer regularizer that affects the predictions of both the task and the evidence labels.

Similarly, we define
\begin{align}
    \sR_2(\theta)
        &= - \E_{x\sim\sD} \pb{
            \ln \underset{k\in \sI_2(\hat{y}(x))}{\max} p(z_k = +1 \mid x ; \theta )  )
        }.
    \label{eq:inconsistency_r2}
\end{align}
Intuitively, $\sR_2(\theta)$ encourages presence of \textit{some} evidence to support predicted task label. Including $\sR_2(\theta)$ in the loss function is equivalent to supplying pseudo positive samples obtained from the predicted task label $\hat{y}$. 

Similar to Equation \ref{eq:inconsistency_r1_soft}, we can penalize insufficiency using posterior probability as weights,
\begin{align}
    \widetilde{\sR}_2(\theta)
        = - \E_{x\sim\sD} \Bigg[ \Bigg.
        \sum_{c\in[C]} & p(y=c\mid x;\theta) \, \cdot  \nonumber \\ 
        & \ln \underset{k\in \sI_2(c)}{\max} p(z_k = +1 \mid x ; \theta )  )
    \Bigg. \Bigg].
    \label{eq:inconsistency_r2_soft}
\end{align}

In our work, we focus on regularizers $\sR_1(\theta)$ and $\sR_2(\theta)$. In Appendix \ref{apd:second}, we provide preliminary comparison of regularizers $\sR_1(\theta)$ and $\sR_2(\theta)$ with the \textit{soft} regularizers $\widetilde{\sR}_1(\theta)$ and $\widetilde{\sR}_2(\theta)$. We leave further investigation of soft regularizers to future work.

\subsection{Optimization}
\label{sec:optimization}

We apply deep multitask learning for joint predictions of $y,z_1,\cdots,z_K$. In particular, we parameterize $p(y\mid x;\theta)$ and $p(z_k\mid x;\theta)$ for $k=1,\ldots,K$ with neural network $f(x;\theta)$ and assume function $f$ outputs logits over $K+1$ marginals. 

Given a classification loss function $L(\cdot,\cdot)$, the objective is simply the empirical risk,
\begin{align}
    \mathcal{L}(\theta)
        &= \E_{(x,y)\sim \sD_t} \pb{
            L(y, f(x;\theta))
        } \\ 
    &+ \frac{1}{K} \sum_{k=1}^K \E_{(x,z_k)\sim \sD_t} \pb{
            L(z_k, f(x;\theta))
    }.
\end{align}

We add consistency regularization to multitask classification loss, which yields a regularized empirical risk minimization problem
\begin{align}
    \min_{\theta}\,
        \mathcal{L}(\theta) + \omega_1 \sR_1(\theta) + \omega_2 \sR_2 (\theta),
\end{align}
where $\omega_1,\omega_2\in\R^+$ are coefficients that control the degree of regularization.

\begin{figure}[!hb]
\floatconts
    {fig:results_ic_v3_measure_inconsistency}
    {\caption{Model inconsistency across different values of $\hat{y}$ evaluated on the test set $\hat{\sD}$ for a naively trained model, i.e., $\omega_1=\omega_2=0$. The majority of inconsistent evidence comes from incompatible evidence associated with small values of the predicted task label $\hat{y}$.}}
    {%
    \includegraphics[width=\linewidth]{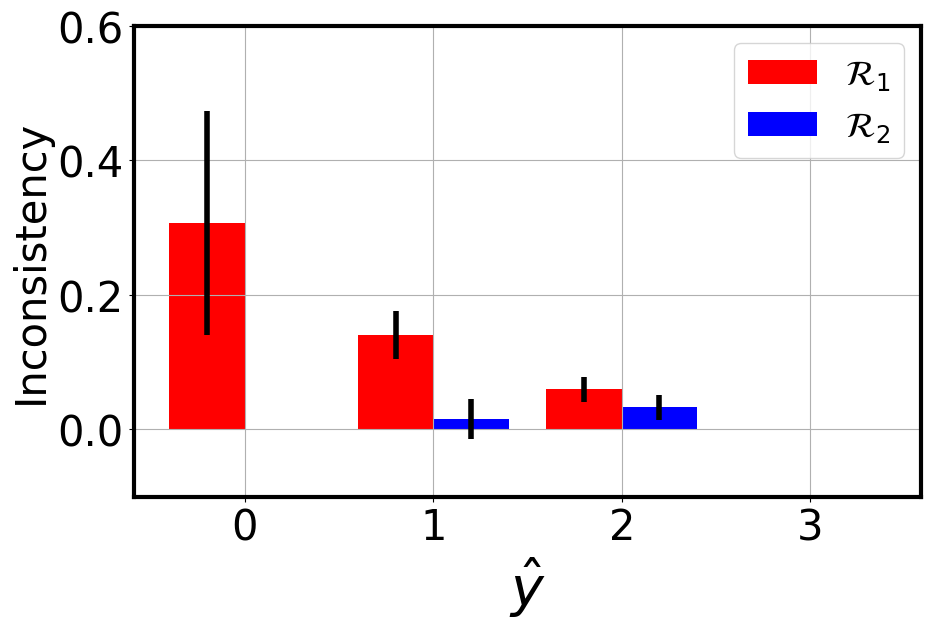}%
    }
\end{figure}

\begin{figure*}[!h]
\floatconts
    {fig:consistentreg_vary_regularization_strength}
    {\caption{The effect of varying strength of regularizations during training on model consistency. Here $\sR_1,\sR_2$ is short hand for $\sR_1(\hat{\sD}),\sR_2(\hat{\sD})$, respectively. The proposed regularizers $\sR_1(\theta),\sR_2(\theta)$ encourage the model to provide more compatible evidence (left) or more sufficient evidence (middle), respectively. The application of regularizers at the same time encourages the model to provide more consistent evidence (right).}}
    {%
    \subfigure[Vary $\omega_1$, fix $\omega_2=0$]{\label{fig:results_ic_v3_consistentreg_vary_ic1}%
        \includegraphics[width=.3\linewidth]{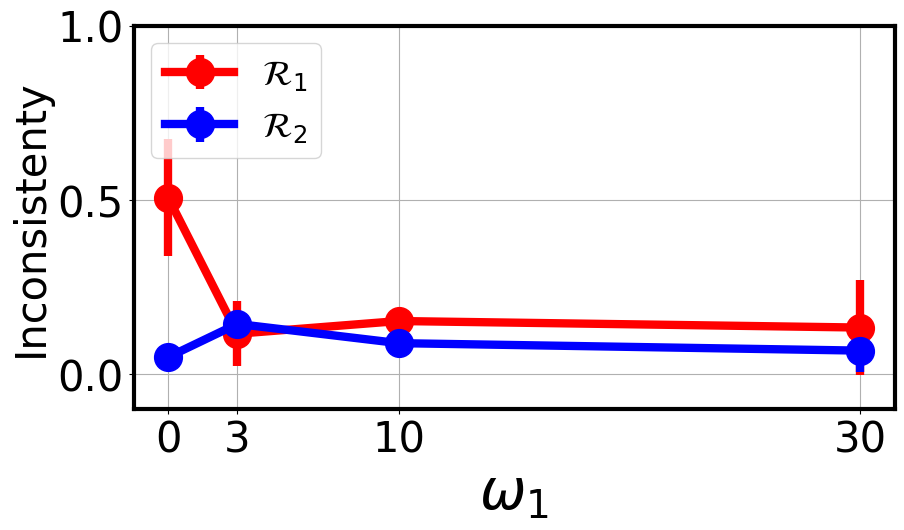}}%
    \qquad
    \subfigure[Fix $\omega_1=0$, vary $\omega_2$]{\label{fig:results_ic_v3_consistentreg_vary_ic2}%
        \includegraphics[width=.3\linewidth]{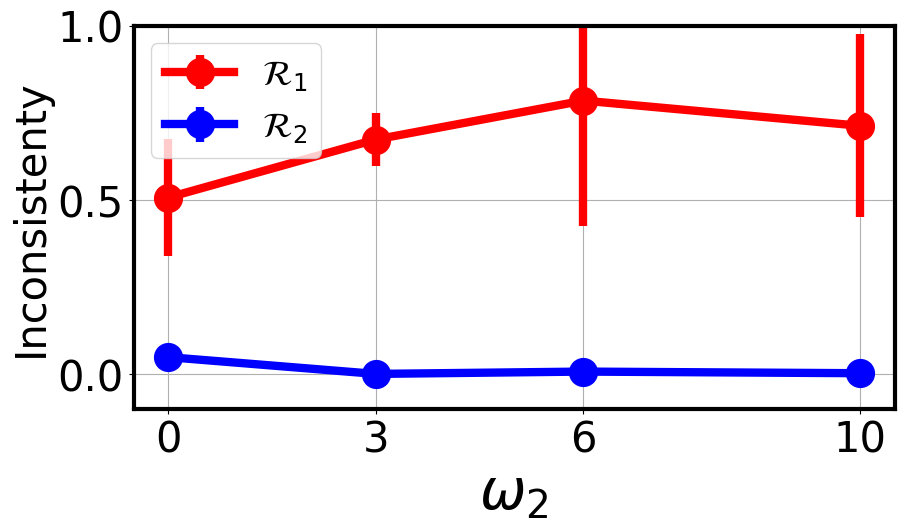}}
    \qquad
    \subfigure[Vary $\omega_1=\omega_2$ together]{\label{fig:results_ic_v3_consistentreg_vary_ic}%
        \includegraphics[width=.3\linewidth]{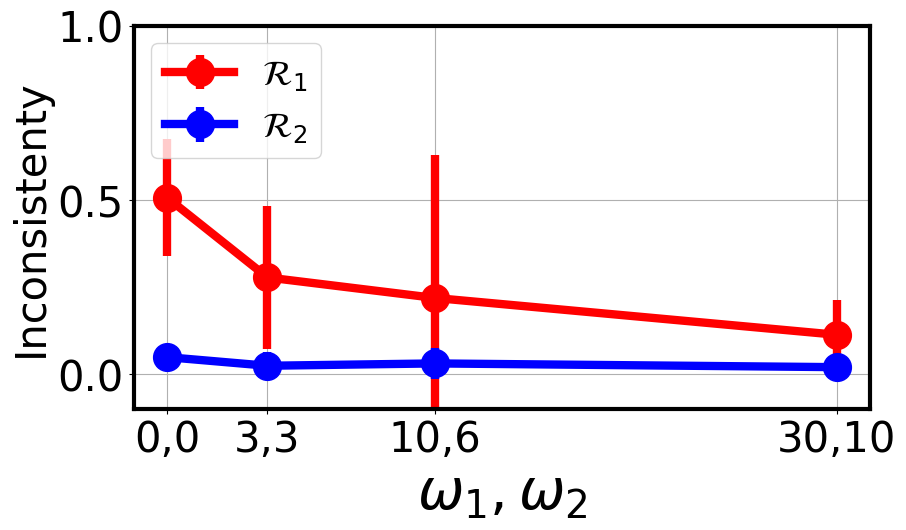}}
    }
\end{figure*} 

\section{Implementation Details} 

We use residual networks to parameterize our probabilistic classifiers \citep{heDeepResidualLearning2016}. The network is modified to output a $(C+K)$-dimensional vector representing the posterior marginal probabilities for $y, z_1,\ldots,z_K$. 

We use weighted cross entropy loss as $L(\cdot,\cdot)$ to handle class imbalances. We employ the Adam optimizer with a constant learning rate of $2\cdot 10^{-4}$ with mini-batch size of 32 for stochastic optimization of network parameters \citep{kingmaAdamMethodStochastic2015}. Each gradient update involves random sampling of a label (task or evidence), assembling a mini-batch of data corresponding to the sampled label, computing the objective function, and updating parameters with backpropagated gradients. This approach enables us to learn even if some labels are missing for some images.

We normalize images to zero mean and unit variance, and resize them to 224x224 pixels. We apply random image augmentations to images, e.g., crop, horizontal flip, brightness and contrast variations, to alleviate model overfitting. 

We implement Equation \ref{eq:inconsistency_r2} by substituting the $\max$ operator with a soft maximum operator, i.e., $\text{LSE}_{i\in[n]}(x_i) = \log\sum_i\exp(x_i)$. This way, we enforce sufficiency of evidence by upscaling probabilities of direct evidence that are larger to begin with.

We use exactly the same network architecture, data augmentation, and optimization parameters to isolate the impact of the proposed regularization on consistency and performance.

We compute mean and standard deviation statistics for inconsistency and test prediction from 3 runs with different random seed.

\section{Experiments}

\subsection{Data}

We use a subset of 238,086 frontal-view chest X-ray from the MIMIC-CXR data set \citep{johnsonMIMICCXRDeidentifiedPublicly2019}. We split the data set into training (217,016), validation (10,445), and test (10,625) sets randomly. The performance of predicted evidence is computed over this test set. There is no patient overlap between training, validation and test sets.

Edema severity labels are extracted from associated reports by searching for keywords that are indicative of a specific disease stage. The 7,802 labeled image/report pairs are split into training (6,656), validation (648), and test (498) set. The test set was corrected for keyword matching errors by an expert radiologist, as detailed in prior work \citep{chauhanJointModelingChest2020}. We use $\hat{\sD}$ to denote this test set that includes images and predicted labels $(x, \hat{y}, \hat{z})$. All subsequent evaluations of model consistency and performance is computed using $\hat{\sD}$.

\subsection{Model Inconsistency}

We examine model inconsistency overall and over partitions of data with respect to values of predicted label $\hat{y}$. The sum of model inconsistency over the partitions gives the quantities $\sR_1(\hat{\sD})$ in Equation \ref{eq:inconsistency_measure_incompatible_upper_bound_dataset} and $\sR_2(\hat{\sD})$ in Equation \ref{eq:inconsistency_measure_insufficiency_upper_bound_dataset}.

Figure \ref{fig:results_ic_v3_measure_inconsistency} reports model inconsistency over partitions of $\hat{y}$ for a model that is trained without consistency regularization, i.e., $\omega_1=\omega_2=0$. We observe that $\sR_1(\hat{\sD})$ is typically larger than $\sR_2(\hat{\sD})$ due to the fact that $\sR_1(\hat{\sD})$ is essentially an average count of potentially many incorrect evidence labels, while $\sR_2(\hat{\sD})$ is an average count of missing evidence, and therefore is upper bounded by 1. We also observe a downward trend in values of $\sR_1(\hat{\sD})$ with increasing values for $\hat{y}$. This is reasonable as there are many ways to make mistake with a small $\hat{y}$, while no way to provide conflicting evidence when $\hat{y}=3$ in our framework.

\begin{figure*}[!ht]
\floatconts
    {fig:test_image_with_evidence}
    {\caption{Correctly and incorrectly classified test images with supporting evidence given by a consistent model ($\omega_1=\omega_2=10.$). We use $\sI(c)$ to denote the set of evidence labels detected in the image that directly support disease stage $c$.}}
    {%
    \includegraphics[width=.5\linewidth]{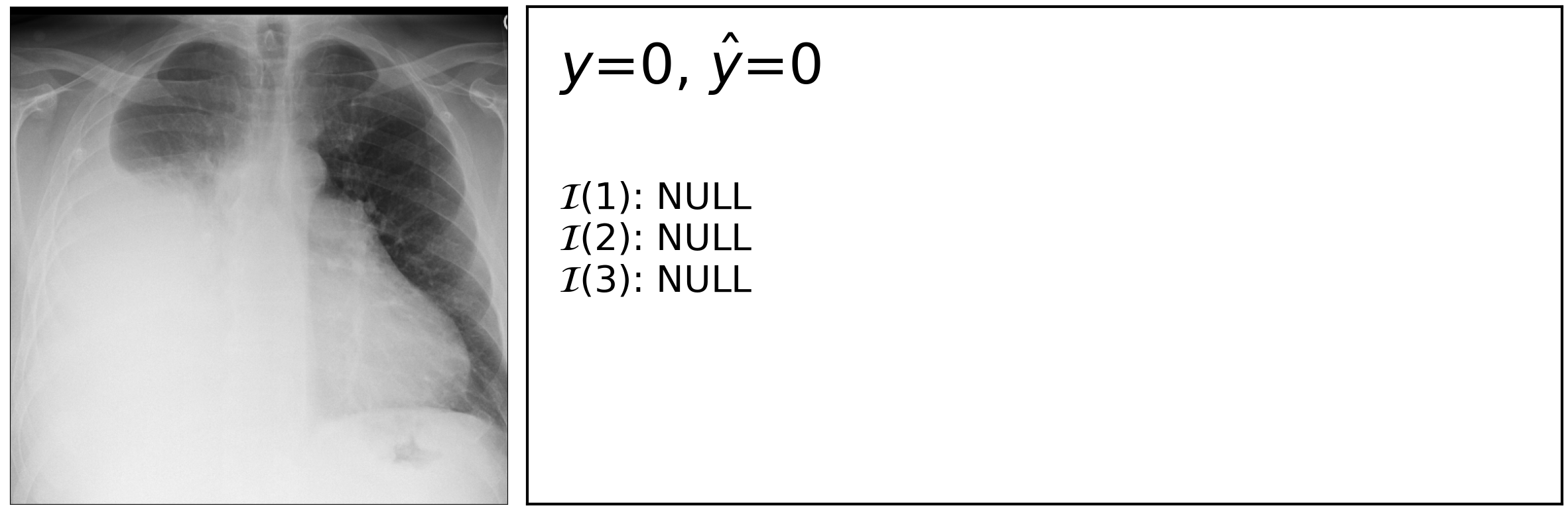}%
    \includegraphics[width=.5\linewidth]{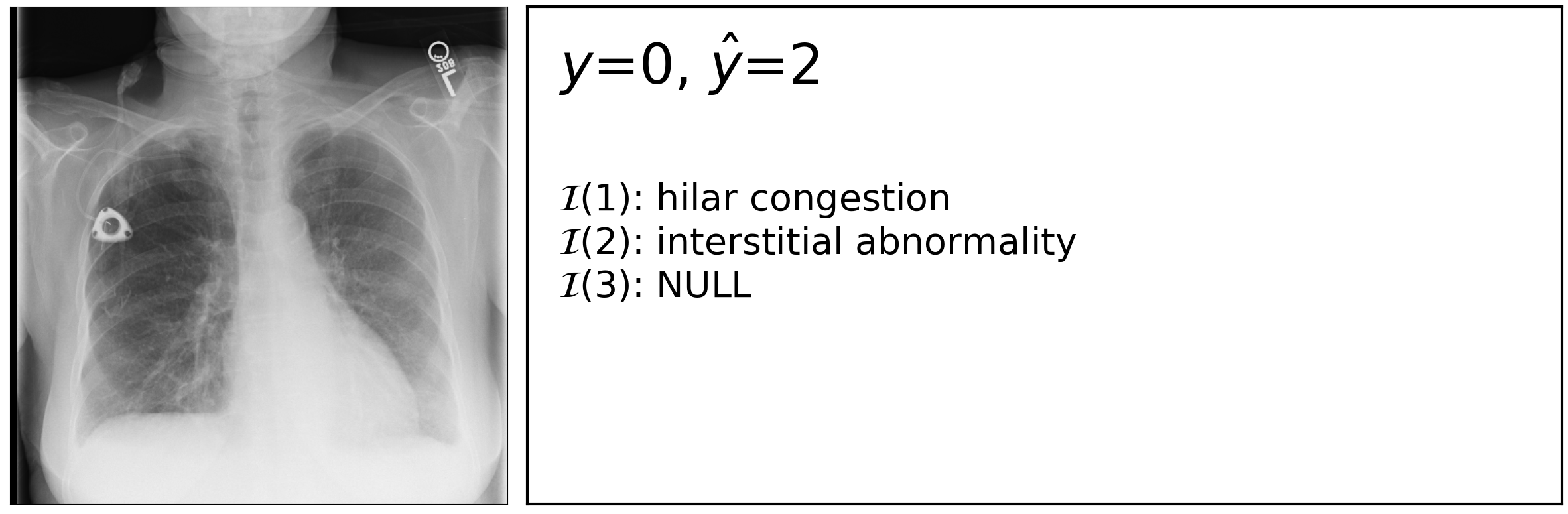}%
    \hfill
    \includegraphics[width=.5\linewidth]{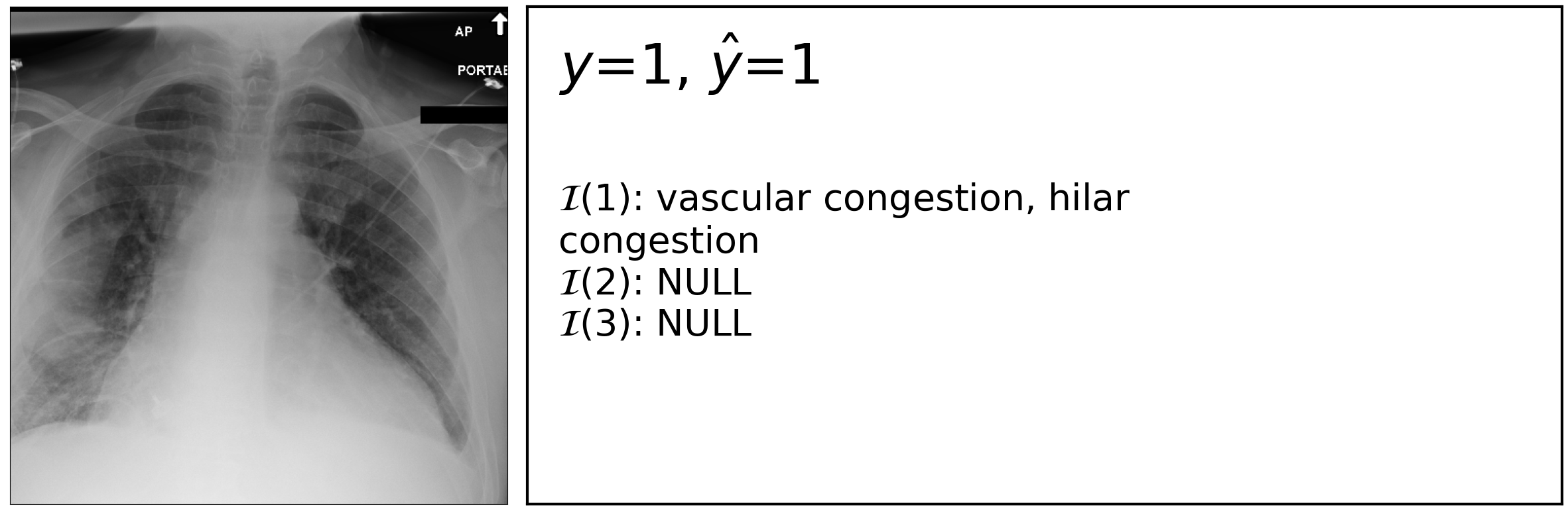}%
    \includegraphics[width=.5\linewidth]{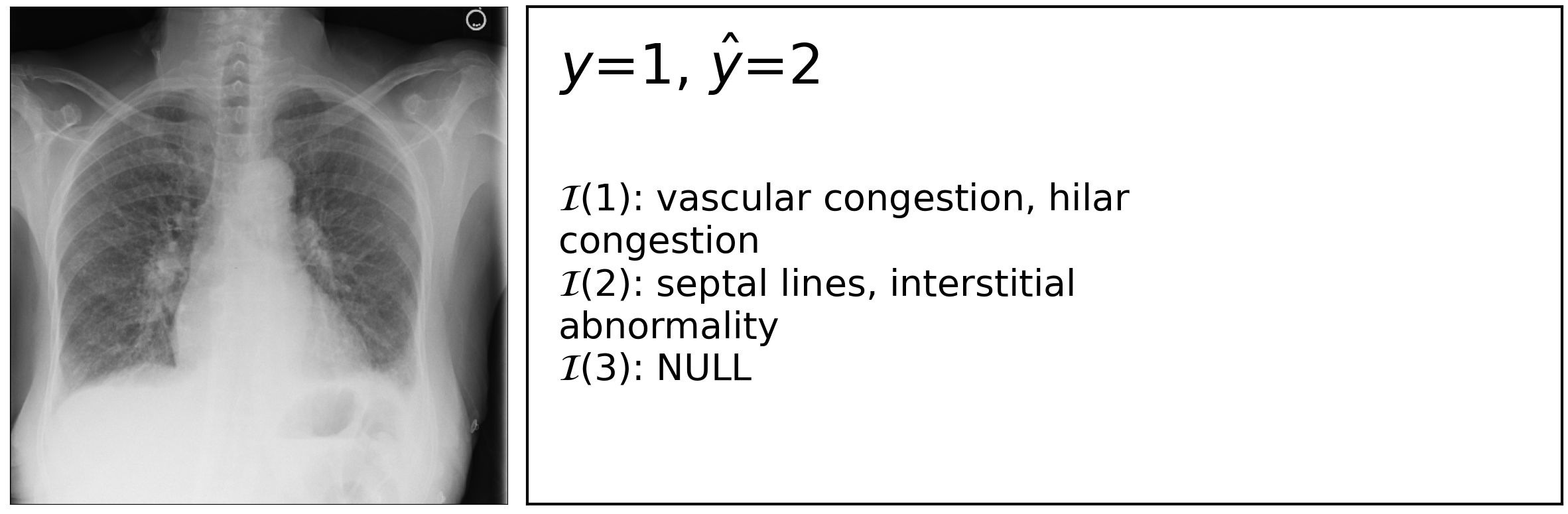}%
    \hfill
    \includegraphics[width=.5\linewidth]{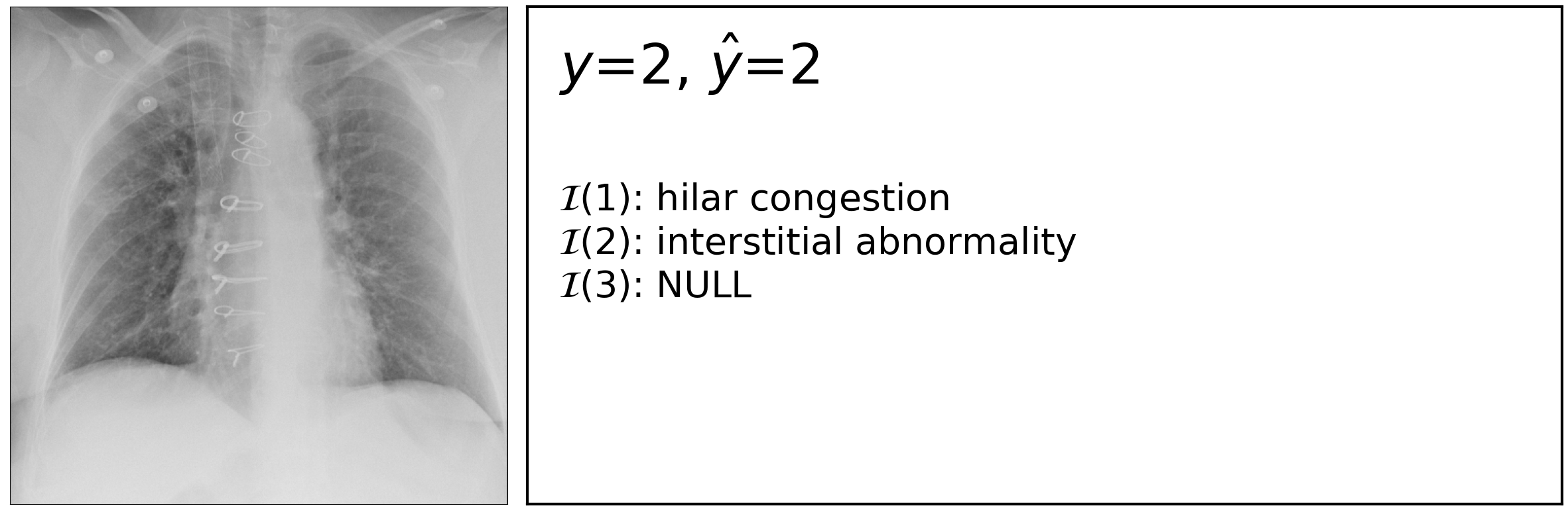}%
    \includegraphics[width=.5\linewidth]{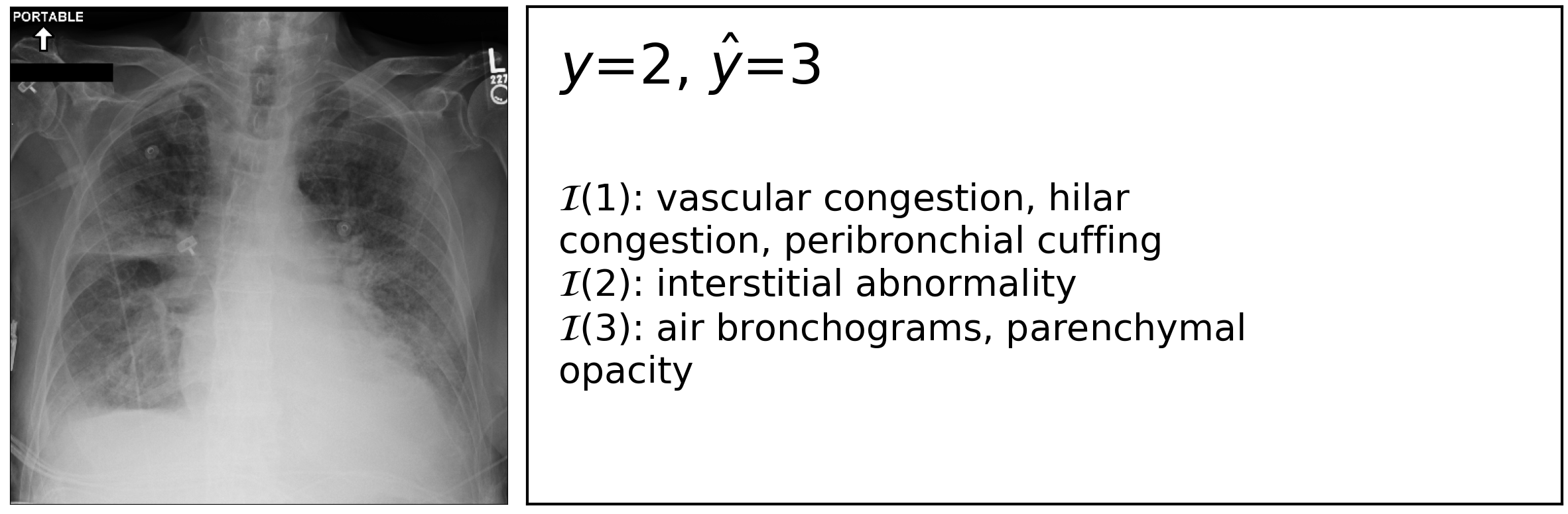}%
    \hfill
    \includegraphics[width=.5\linewidth]{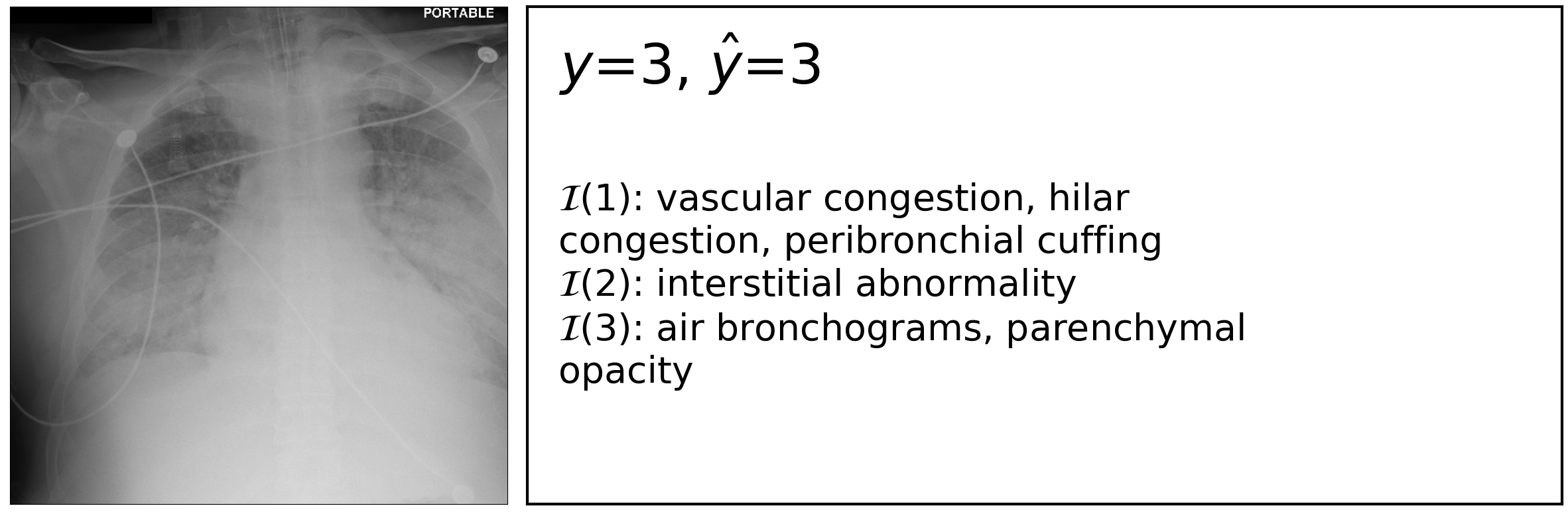}%
    \includegraphics[width=.5\linewidth]{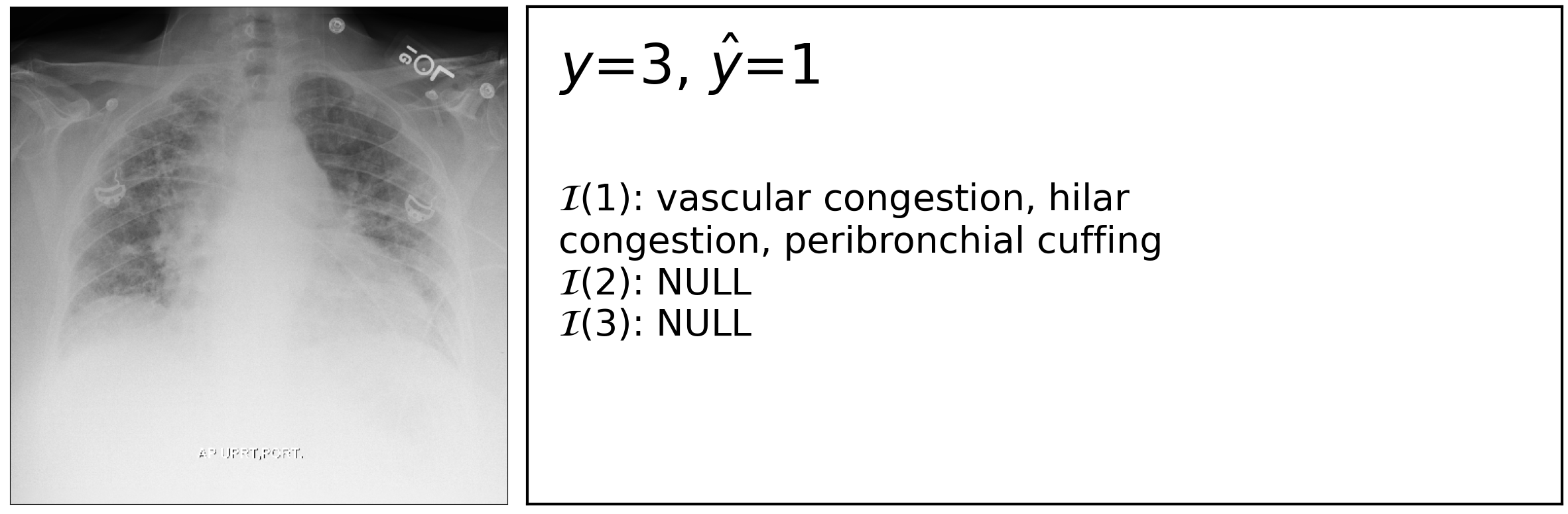}%
    }
\end{figure*}

\subsection{Consistency Regularization}

To demonstrate that proposed regularization promotes model consistency, we vary values of $\omega_1,\omega_2$ in the objective function and train multiple models. We select the most accurate model on the validation set and compute inconsistency on the test set $\hat{\sD}$.

Figure \ref{fig:consistentreg_vary_regularization_strength} demonstrates the effects of regularization on model consistency. We observe that the regularizers $\sR_1(\theta)$ and $\sR_2(\theta)$ are effective in reducing the respective intended model inconsistency, indicated by a reduction of $\sR_1(\hat{\sD})$ in Figure \ref{fig:results_ic_v3_consistentreg_vary_ic1} and $\sR_2(\hat{\sD})$ in Figure \ref{fig:results_ic_v3_consistentreg_vary_ic2} respectively. Additionally, we observe that penalizing $\sR_1(\hat{\sD})$ inadvertently makes $\sR_2(\hat{\sD})$ larger and vice versa. This makes intuitive sense, since a model that is more likely to predicts absence of evidence will (i) less likely to provide incompatible evidence and (ii) less likely to provide some direct evidence. We observe that we can reduce both types of inconsistency by regularizing with both loss terms, as shown in Figure \ref{fig:results_ic_v3_consistentreg_vary_ic}. 

It important to note that even though $\sR_2(\hat{\sD})$ is relatively small in models trained with $\omega_1=0$, regularizing with $\sR_2(\hat{\sD})$ is necessary as we want to avoid situations in Figure \ref{fig:results_ic_v3_consistentreg_vary_ic1} where $\sR_2(\hat{\sD})$ becomes intolerably large.

\subsection{Interpretability}

\begin{figure*}[!ht]
\floatconts
    {fig:perf_consist_tradeoff}
    {\caption{The effect of regularization on model inconsistency (left 2) and performance of predicting task label $y$ (middle 2) and evidence labels $z$ (right 2). Here $\sR_1,\sR_2$ is short hand for $\sR_1(\hat{\sD}),\sR_2(\hat{\sD})$, respectively. When regularizing for both $\sR_1,\sR_2$, e.g., down the diagonals of the matrix, we notice dramatic decrease in model inconsistency and competitive performance for the regularized model. }}
    {%
    \includegraphics[width=\linewidth]{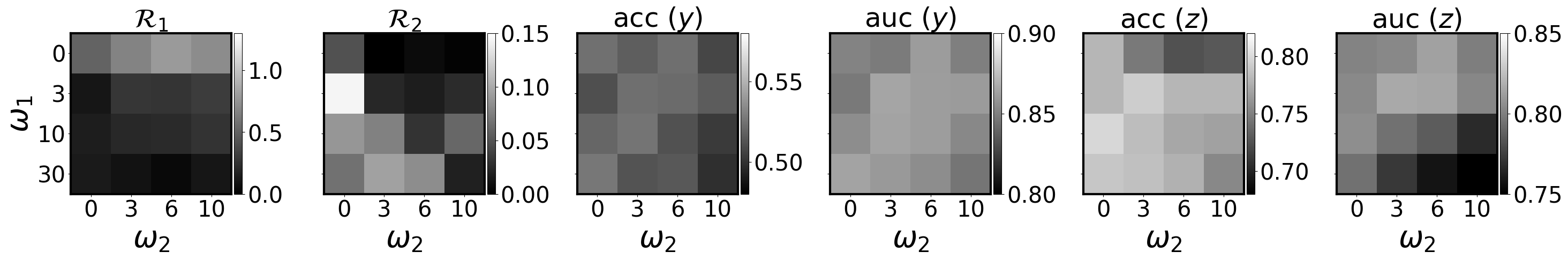}%
    }
\end{figure*}

Figure \ref{fig:test_image_with_evidence} illustrates how a consistent model (trained with $\omega_1=\omega_2=10$) provides supporting evidence for randomly sampled test images. We provide correctly and incorrectly classified test images for each severity level. We observe that the regularized model provides consistent evidence in all 8 examples, even in cases where model prediction of task label is not correct.

How does providing consistent supporting evidence build trust in the model ? We note that supporting findings are already described in radiological reports and can be easily mined for training and verified in an image. When the supporting evidence is clearly correct, it builds additional trust in the predicted task label. When the consistent but wrong evidence is presented, it is easy to see in the image and helps the end users understand why the main task label is wrong. Our method avoid confusions that arise from model providing inconsistent evidence. 

Crucially, evidence labels should not only be consistent, but also correct. To this point, we reported performance of evidence detection in Section \ref{sec:perf_consistency_tradeoff}. Our proposed regularizers offers a complementary tool to help end users understand why the model erred. Our method can be integrated with technologies, e.g., RCNN, GradCAM \citep{girshickRichFeatureHierarchies2014,selvarajuGradCAMVisualExplanations2016}, that provide localization, i.e., confirmation that the model is focusing on the correct regions in the image.

\subsection{Performance-Consistency Tradeoff}
\label{sec:perf_consistency_tradeoff}

Next, we show that we can achieve good model consistency without compromising predictive performance. We vary $\omega_1,\omega_2$ together in the objective function and train multiple models. We select for the most accurate model on the validation set for subsequent evaluations.

Figure \ref{fig:perf_consist_tradeoff} demonstrates that we can ensure satisfactory model consistency. At the same time, the regularized model achieves similar performance on the severity grading task. The improvement in performance can be attributed empirically to fact that heavily regularized models over-fit less. However, we do observe a drop in the average performance of the model for predicting evidence. The drop in predicted evidence performance is tolerable if we consider that the model rarely provides inconsistent supporting evidence.

Figure \ref{fig:perf_consist_tradeoff} reinforces previous observation that penalizing $\sR_1(\hat{\sD})$ makes $\sR_2(\hat{\sD})$ higher and vice versa, when $\omega_i$ for $i=1,2$ being held constant is different from 0. Figure \ref{fig:consistentreg_vary_regularization_strength} shows the first column, the first row, and the diagonal slices of the grid in left 2 sub-figures in Figure \ref{fig:perf_consist_tradeoff}. We refer the reader to Table \ref{tab:consistency_performance_tradeoff} in Appendix \ref{apd:first} for detailed numbers of inconsistency and performance along the diagonal slice of the grid.

\section{Conclusion}

We argue for supplementing model predictions with supporting evidence that is deemed useful by end users. We defined a notion of consistent evidence via incorporating domain specific constraints. Then, we proposed ways to measure and enforce such constraints during model training. We evaluated our method on the pulmonary edema severity grading task, which provides a grounding for our consistent evidence framework. We demonstrated that consistent models remain competitive on the main task.  

\acks{This work was supported in part by NIH NIBIB NAC P41EB015902 grant, MIT Lincoln Laboratory, MIT JClinic, MIT Deshpande Center, and Philips.}

\newpage

\bibliography{multilabel, interpretability, cxr, ml4h, neurosymbolic, network_vision, opt_nn}

\appendix

\section{Table for Figure 5.}\label{apd:first}

Table \ref{tab:consistency_performance_tradeoff} provides detailed numbers on inconsistency and performance.

\begin{table*}[!h]
    \floatconts
    {tab:consistency_performance_tradeoff}
    {\caption{The effect of regularization on model inconsistency and performance of predicting task label $y$ and evidence label $z$. $*$ stands for baseline model that is trained to predict task label $y$. Standard deviation is shown after $\pm$. We notice dramatic decrease in model inconsistency and competitive performance for the regularized model.}}
    {
        \begin{tabular}{llllll}
\toprule
               $\omega_1,\omega_2$ &                                    * &                              0.0,0.0 &                              3.0,3.0 &                             10.0,6.0 &                            30.0,10.0 \\
\midrule
$\mathcal{R}_1(\hat{\mathcal{D}})$ &                                    - & $0.507\pm\text{\footnotesize 0.084}$ & $0.277\pm\text{\footnotesize 0.103}$ & $0.218\pm\text{\footnotesize 0.205}$ & $0.112\pm\text{\footnotesize 0.049}$ \\
$\mathcal{R}_2(\hat{\mathcal{D}})$ &                                    - & $0.048\pm\text{\footnotesize 0.011}$ & $0.023\pm\text{\footnotesize 0.019}$ & $0.030\pm\text{\footnotesize 0.023}$ & $0.019\pm\text{\footnotesize 0.007}$ \\
\midrule
                         acc ($y$) & $0.524\pm\text{\footnotesize 0.017}$ & $0.524\pm\text{\footnotesize 0.005}$ & $0.523\pm\text{\footnotesize 0.008}$ & $0.512\pm\text{\footnotesize 0.014}$ & $0.499\pm\text{\footnotesize 0.022}$ \\
                         auc ($y$) & $0.836\pm\text{\footnotesize 0.008}$ & $0.852\pm\text{\footnotesize 0.004}$ & $0.865\pm\text{\footnotesize 0.004}$ & $0.862\pm\text{\footnotesize 0.002}$ & $0.846\pm\text{\footnotesize 0.006}$ \\
\midrule
         acc (vascular congestion) &                                    - & $0.795\pm\text{\footnotesize 0.018}$ & $0.793\pm\text{\footnotesize 0.026}$ & $0.777\pm\text{\footnotesize 0.005}$ & $0.763\pm\text{\footnotesize 0.015}$ \\
            acc (hilar congestion) &                                    - & $0.771\pm\text{\footnotesize 0.047}$ & $0.801\pm\text{\footnotesize 0.031}$ & $0.759\pm\text{\footnotesize 0.016}$ & $0.679\pm\text{\footnotesize 0.040}$ \\
       acc (peribronchial cuffing) &                                    - & $0.804\pm\text{\footnotesize 0.041}$ & $0.802\pm\text{\footnotesize 0.032}$ & $0.795\pm\text{\footnotesize 0.026}$ & $0.789\pm\text{\footnotesize 0.040}$ \\
                acc (septal lines) &                                    - & $0.869\pm\text{\footnotesize 0.010}$ & $0.859\pm\text{\footnotesize 0.043}$ & $0.796\pm\text{\footnotesize 0.037}$ & $0.845\pm\text{\footnotesize 0.023}$ \\
    acc (interstitial abnormality) &                                    - & $0.649\pm\text{\footnotesize 0.011}$ & $0.647\pm\text{\footnotesize 0.002}$ & $0.630\pm\text{\footnotesize 0.014}$ & $0.606\pm\text{\footnotesize 0.006}$ \\
            acc (air bronchograms) &                                    - & $0.863\pm\text{\footnotesize 0.019}$ & $0.881\pm\text{\footnotesize 0.030}$ & $0.888\pm\text{\footnotesize 0.015}$ & $0.860\pm\text{\footnotesize 0.018}$ \\
         acc (parenchymal opacity) &                                    - & $0.706\pm\text{\footnotesize 0.018}$ & $0.764\pm\text{\footnotesize 0.020}$ & $0.756\pm\text{\footnotesize 0.009}$ & $0.738\pm\text{\footnotesize 0.009}$ \\
\bottomrule
\end{tabular}
    }
\end{table*}

\section{Training with Soft Regularizer}\label{apd:second}

We compare behavior of model trained using hard regularizers $\sR_1(\theta),\sR_2(\theta)$ versus the soft regularizers $\tilde{\sR}_1(\theta),\tilde{\sR}_2(\theta)$.

In Figure \ref{fig:perf_consist_tradeoff_compare_hard_soft}, we note that the soft regularizers reduce model inconsistency while maintain model consistency in a similar manner to hard regularizers. Unlike those trained with hard regularizers, models trained with soft regularizers seem to avoid the decrease in performance for evidence detection.

\begin{figure*}[!ht]
\floatconts
    {fig:perf_consist_tradeoff_compare_hard_soft}
    {\caption{The effect of regularization on model inconsistency (left 2) and performance of predicting task label $y$ (middle 2) and evidence labels $z$ (right 2), when model is trained using the hard regularizers $\sR_1(\theta),\sR_2(\theta)$ (top) versus the soft regularizers $\tilde{\sR}_1(\theta),\tilde{\sR}_2(\theta)$ (bottom). Here $\sR_1,\sR_2$ is short hand for $\sR_1(\hat{\sD}),\sR_2(\hat{\sD})$, respectively. }}
    {%
    \includegraphics[width=\linewidth]{images/results_ic_v3_perf_consist_tradeoff_matrix}
    \includegraphics[width=\linewidth]{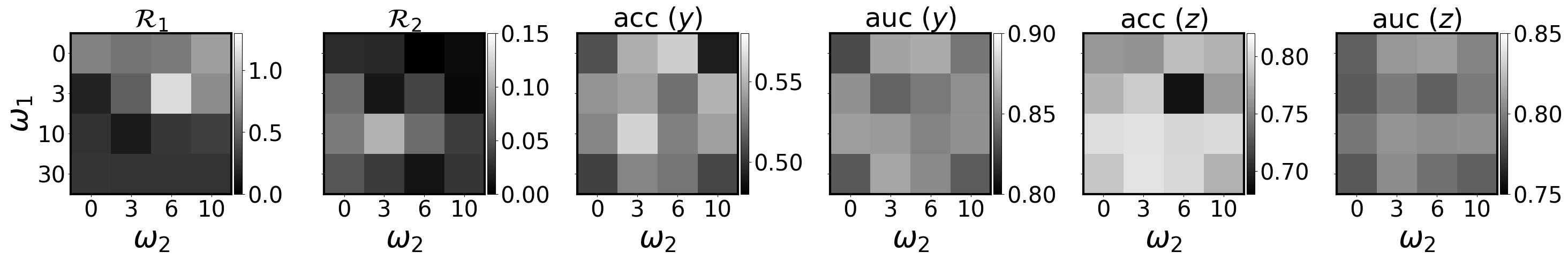}%
    }
\end{figure*}

\end{document}